\documentclass[10pt,twocolumn,letterpaper]{article}

\usepackage[pagenumbers]{cvpr} 

\usepackage[dvipsnames]{xcolor}

\definecolor{cvprblue}{rgb}{0.21,0.49,0.74}
\usepackage[pagebackref,breaklinks,colorlinks,citecolor=cvprblue]{hyperref}
\usepackage{mathtools}
\usepackage{gensymb}

\def\paperID{*****} 
\def\confName{CVPR}
\def\confYear{2024}

\title{Enhancing Polygonal Building Segmentation via Oriented Corners}

\author{Mohammad Moein Sheikholeslami$^1$, Muhammad Kamran$^1$, Andreas Wichmann$^2$, and Gunho Sohn$^{1}$\thanks{Corresponding author.}\\
$^1$Department of Earth and Space Science and Engineering, York University, Canada\\
$^2$Institute for Applied Photogrammetry and Geoinformatics, Jade University, Germany\\
{\tt\small \{mmoein, mkamran9, gshon\}@yorku.ca, andreas.wichmann@jade-hs.de}
}

\begin{document}
\maketitle
\begin{abstract}
The growing demand for high-resolution maps across various applications has underscored the necessity of accurately segmenting building vectors from overhead imagery. However, current deep neural networks often produce raster data outputs, leading to the need for extensive post-processing that compromises the fidelity, regularity, and simplicity of building representations. In response, this paper introduces a novel deep convolutional neural network named OriCornerNet, which directly extracts delineated building polygons from input images. Specifically, our approach involves a deep model that predicts building footprint masks, corners, and orientation vectors that indicate directions toward adjacent corners. These predictions are then used to reconstruct an initial polygon, followed by iterative refinement using a graph convolutional network that leverages semantic and geometric features. Our method inherently generates simplified polygons by initializing the refinement process with predicted corners. Also, including geometric information from oriented corners contributes to producing more regular and accurate results.
Performance evaluations conducted on SpaceNet Vegas and CrowdAI-small datasets demonstrate the competitive efficacy of our approach compared to the state-of-the-art in building segmentation from overhead imagery.
\end{abstract}    
\section{Introduction}
\label{sec:intro}
\begin{figure}[t]
  \centering
   \includegraphics[width=0.9\linewidth]{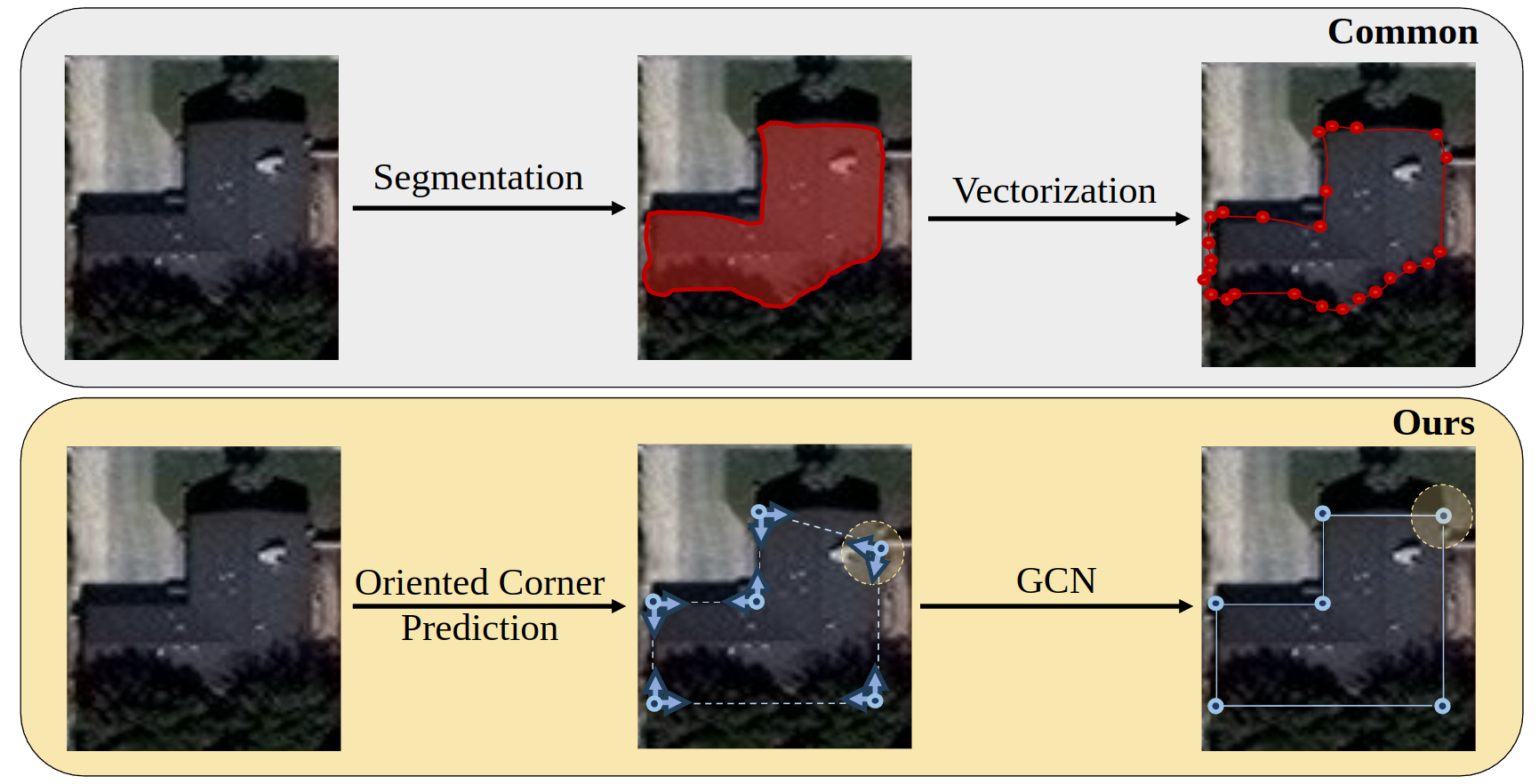}

   \caption{Comparison between the common approach for extracting buildings from imagery and OriCornerNet. The common approach suffers from redundant vertices and irregular polygons.}
   \label{fig:pipeline}
\end{figure}

The integration of high-resolution satellite imagery and unmanned aerial vehicle (UAV) data has had a profound impact on sectors like urban planning and monitoring of the built environment. Despite these advancements, the creation of accurate maps remains a resource-intensive endeavor, particularly in the realm of building extraction. This process traditionally comprises two key phases: firstly, the segmentation of building footprints from imagery, and secondly, the conversion of these footprints into vector formats compatible with geographic information systems (GIS). Challenges such as shadows, tree obstructions, and errors in raster quantization often lead to imperfect masks during segmentation, resulting in suboptimal polygons during the vectorization process (see~\Cref{fig:pipeline}).

Within the domain of deep learning methods for polygonal building segmentation, two primary categories have emerged: 1) two-step methodologies and 2) direct polygonal segmentation approaches~\cite{dominici2020polyfit, asip}. Two-step methods, akin to conventional techniques, involve initial raster segmentation followed by post-processing steps, including vectorization and potential simplification. They frequently employ auxiliary representations such as frame fields~\cite{ff}, directional indicators~\cite{robust}, or attraction fields~\cite{hisup} to aid in the vectorization process. However, these methods are constrained by their reliance on offline post-processing, which can fail under certain conditions. In contrast, direct polygonal segmentation strategies directly predict building polygons from input images, sidestepping the challenges associated with two-step methodologies. Nevertheless, these methods can be intricate to train, computationally demanding, and may encounter difficulties such as irregular predictions or missing corners~\cite{polymapper, rpolygcn, polyworld}.

\begin{figure*}[t]
  \centering
   \includegraphics[width=0.9\linewidth]{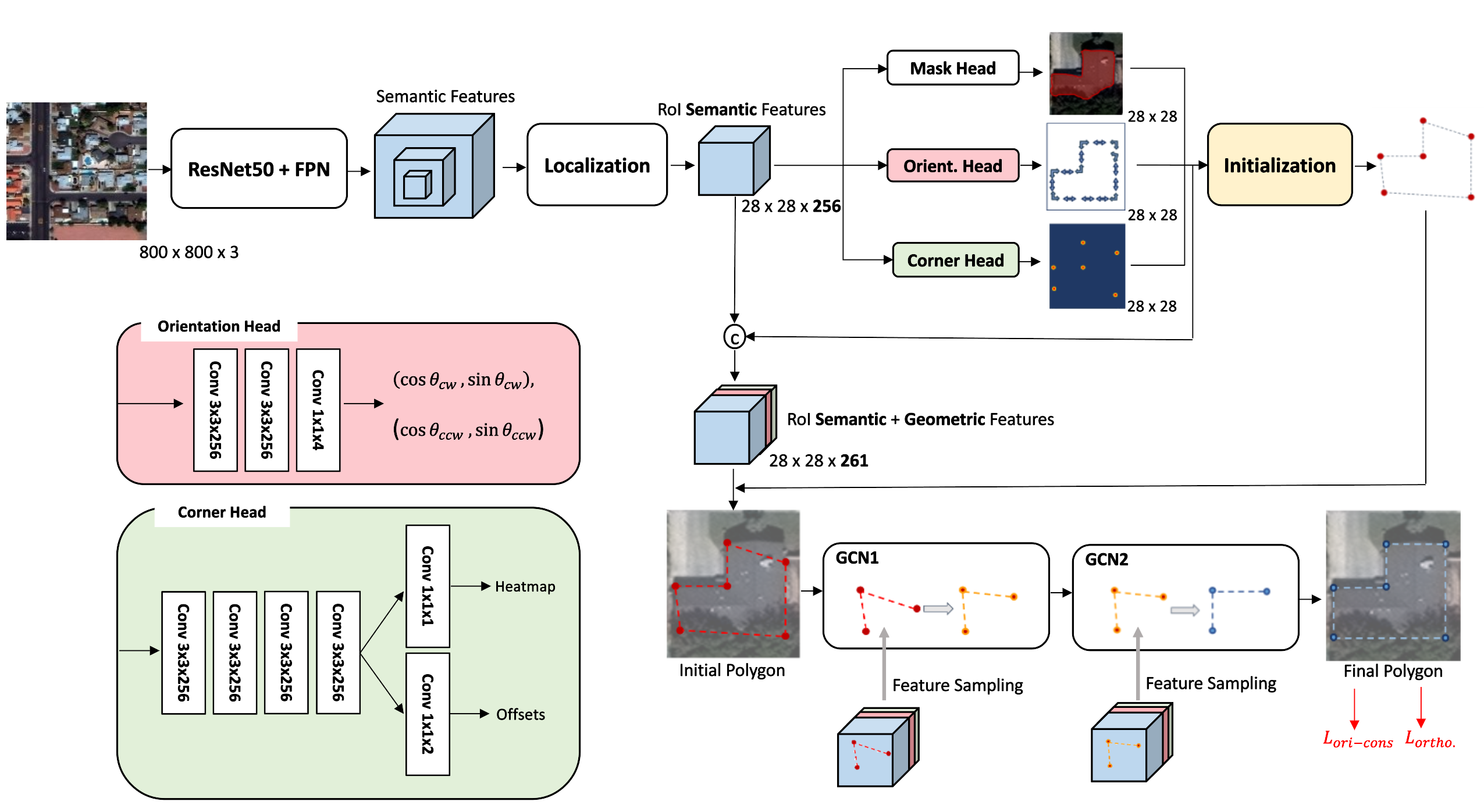}

   \caption{The architecture of OriCornerNet.}
   \label{fig:arch}
\end{figure*}

This paper presents a novel deep neural network named OriCornerNet, designed to enhance the performance of a baseline direct polygonal segmentation model called R-PolyGCN~\cite{rpolygcn}. OriCornerNet achieves this by integrating the detected oriented corners as an auxiliary representation. This innovative approach, similar to direct segmentation methods, eliminates the need for post-processing steps like vectorization. Also, it utilizes the auxiliary representation to guide the generation of building polygons like two-step methodologies. In this context, 'orientation' refers to vectors associated with each corner that point towards adjacent corners. Our method uses auxiliary representation to generate regular polygons, identifying occluded corners and capturing the architectural nuances effectively.

\section{Related Work}
In this section, we will briefly review polygonal segmentation methods that utilize auxiliary information to guide them through the vectorization process. These methods are mostly two-step approaches. For example, Girard \etal~\cite{ff} predict a raster segmentation along with frame fields and use the latter to vectorize the predicted raster segmentation. Similarly, Shu \etal~\cite{robust} predict corners and direction vectors using a deep network and then connect the corners by a boundary tracing algorithm. Another approach adopted by \cite{zhu2021adaptive} formulates the polygons as a combination of vertex and direction maps predicted by deep networks. Then, a polygon generation algorithm is used to generate the final output. Recently, HiSup~\cite{hisup} proposed to use the attraction field as a mid-level supervision in mask and corner segmentation tasks, leading to a state-of-the-art performance in polygonal building segmentation.

The major disadvantage of the methods mentioned earlier is requiring rule-based post-processing, which is susceptible to failure. Additionally, two-step methods experience significant issues with raster quantization error, as noted in \cite{ff}. This emphasizes the need for direct methods to effectively use auxiliary information.
\section{Methodology}
For our model, we selected R-PolyGCN~\cite{rpolygcn} as the baseline, which comprises three main stages: an instance segmentation model, graph initialization, and graph convolutional networks (GCNs). Then, we modified each stage to improve polygonal segmentation performance using oriented corners, as shown in~\Cref{fig:arch} and described below.

\subsection{Instance Segmentation Baseline}
Following R-PolyGCN~\cite{rpolygcn}, we use Mask R-CNN~\cite{maskrcnn} as the baseline for semantic mask segmentation. In addition, we add corner and orientation heads. Given an input image $I\in \mathbb{R}^{800\times800\times3}$, semantic features $S\in \mathbb{R}^{28\times28\times256}$ are extracted for each RoI and inputs to the heads, then the corner head, fed by the $S$, predicts a heatmap $H \in \mathbb{R}^{28\times28}$ and xy offset maps $\Delta\in\mathbb{R}^{28\times28\times2}$. To be concise, we only bring the losses for corner and orientation heads:
{\small
\begin{equation} 
\label{eq:heatloss}
L_{\text{heat.}} = -\frac{1}{N} \sum_{i=1}^{N} (w.y_i.\log x_i + (1-y_i) \log(1-x_i))
\end{equation}
}
and
\begin{equation} \label{eq:offsetloss}
\begin{split}
& L_{\text{offset}} = \frac{1}{N} \sum_{i=1}^{N} L_i, \\
& L_i =
\begin{cases}
0.5 (x_i - y_i)^2, & \text{if } |x_i - y_i| < 1 \\
|x_i - y_i| - 0.5, & \text{otherwise}
\end{cases}
\end{split}
\end{equation}
where $x_i$, $y_i$, $w_p$, and $N$ are, respectively, prediction, target, weight of the positive class, and number of pixels in each task.
For orientation representation on each edge pixel, we use two unit vectors $O_{cw}=(\cos{\theta_{cw}}, \sin{\theta_{cw}})$ and $O_{ccw}=(\cos{\theta_{ccw}}, \sin{\theta_{ccw}})$ pointing to the next corner in the polygon in respectively clockwise and counter-clockwise directions. Thus, four channels are regressed and supervised on ground truth edge pixels using the below loss:
\begin{equation} \label{eq:oriloss}
\begin{split}
& L_{\text{orient.}} = \frac{1}{N_e} \sum_{i=1}^{N_e} L_i, \\
& L_i =
\begin{cases}
0.5 (x_i - y_i)^2, & \text{if } |x_i - y_i| < 1 \\
|x_i - y_i| - 0.5, & \text{otherwise}
\end{cases}
\end{split}
\end{equation}
 where $x_i$, $y_i$, and $N_e$ are prediction, target, and the number of edge pixels.

\subsection{Initialization Module}
Many methods in literature~\cite{rpolygcn, dance, curvegcn, deepsnake} use a fixed number of predefined corners, e.g., located on the perimeter of a circle, for creating the initial graph. However, we use the predicted corners and connect them based on their distances from the mask contour. Also, we propose to combine semantic corners (mask's contour points) with the detected ones to enhance the graph initialization accuracy. In this sense, predicted corners further than $\delta_{cor2cont}$ from the contour are removed. Moreover, the semantic corners further than $\delta_{sem2graph}$ from the graph constructed using only detected corners are added to the corners. Lastly, predicted orientations and corner heatmaps, i.e. geometric features $G \in \mathbb{R}^{28\times28\times5}$, are concatenated with RoI semantic features $S$ to construct $F=concat\{S, G\} \in \mathbb{R}^{28\times28\times261}$.
\subsection{GCNs}
This part includes three GCN modules as described in~\cite{curvegcn}. Getting the initial graph, GCN modules repeatedly sample the feature vectors for each corner from $F$ and predict their position offsets. As the number of corners in prediction and target may vary, a bi-projection~\cite{biprojection} loss is used:
{\small
\begin{equation} \label{eq:poly}
L_{\text{poly}} = \frac{1}{N} (\sum_{(p,\bar{p}) \in S_{ic}} \lVert p - \bar{p} \rVert_2 + 
\sum_{(q,\bar{q}) \in S_{ac}} \lVert q - \bar{q} \rVert_2)
\end{equation}
}
where $S_{ic}$ represents the set of initially matched corners based on their distance, and $S_{ac}$ denotes the set of additional corners matched to their projection on the nearest edge. Additionally, $N$ stands for the number of corners in the polygon with a higher count. We also incorporate two regularization losses: 1) orientation consistency loss and 2) orthogonality loss. Orientation consistency loss penalizes the error between predicted orientation vectors in each edge:

\begin{equation}\label{eq:orientcons}
L_{\text{ori-cons}}(P) = \frac{1}{n} \sum_{i=1}^{n} (1+\langle O_{ccw}^{i}, O_{cw}^{i+1} \rangle)
\end{equation}

where $n$ is the number of vertices in polygon $P$ and $\langle ., .\rangle$ is the inner product of the vectors. The idea is that the orientation vectors of two consecutive corners should point to each other.
On the other hand, orthogonality loss encourages all the inner angles to be orthogonal, as is frequently seen in buildings:

\begin{equation} \label{eq:ortholoss}
L_{\text{ortho.}}(P) = \frac{1}{n} \sum_{i=1}^{n} \min_{\omega_{peak}}{\mathopen |\omega_i - \omega_{peak}\mathclose |}
\end{equation}

where $n$ is the number of inner angles in polygon $P$, $\omega$ is the inner angle and $\omega_{peak} \in \{0^{\circ}, 90^{\circ}, 180^{\circ}, 270^{\circ}\}$. The final loss is a weighted sum of all the losses.

\section{Experiments}
\subsection{Datasets and Settings}
The proposed network's performance is evaluated using the SpaceNet Vegas dataset~\cite{van1807spacenet} to demonstrate its superiority over R-PolyGCN, its baseline, and several competing methods. The dataset contains 3,851 images, with a size of 650$\times$650 pixels and a pixel size of 0.3 meter. The dataset is randomly divided into training, validation, and test sets. We trained the model using a single NVIDIA GeForce RTX 3090 GPU with a batch size of 1 for 50 epochs.

Additionally, the CrowdAI-small dataset~\cite{crowdai} is utilized for both ablation studies and comparison with state-of-the-art methods. This dataset consists of 10,186 images sized at 300$\times$300 pixels, with a pixel size of 0.3 meter. Training on the CrowdAI-small dataset is performed for 40 epochs on a single NVIDIA GeForce RTX 6000 GPU with a batch size of 1. Images from both datasets are resized to 800$\times$800 pixels for input to the network. Also, $\delta_{cor2cont}=5$ and $\delta_{sem2graph}=5$ pixels for both datasets. 
\subsection{Evaluation Metrics}
Two groups of metrics have been utilized in the performance assessment of the proposed method: 1) Raster-based MS-COCO metrics~\cite{coco}, such as mean average precision (AP) and recall (AR) over different IoU thresholds, and 2) Vector-based metrics, including PoLiS~\cite{polis} and C-IoU~\cite{hisup}. PoLiS calculates the average distance between the matched vertices in polygons $P$ and $Q$:
{\small
\begin{equation} \label{polis}
\begin{aligned}
PoLiS(P,Q)=\frac{1}{2N_P}\sum_{p_j\in P} \min_{q \in Q} \lVert p_j - q \rVert +\\
\frac{1}{2N_Q}\sum_{q_k\in Q} \min_{p \in P} \lVert q_k - b \rVert,
\end{aligned}
\end{equation}
}
where $N_P$ and $N_Q$ are, respectively, the number of vertices in the predicted and ground truth polygons.
Furthermore, C-IoU~\cite{hisup} measures the polygon complexity as well as its IoU:
\begin{equation} \label{ciou}
\begin{split}
& {CIoU}(P,Q)=IoU(P,Q).(1-D(N_P,N_Q)),
\end{split}
\end{equation}
where $D(N_P,N_Q)=|N_P-N_Q|/(N_P+N_Q)$ is the relative difference between the number of vertices in the polygons.
\subsection{Preliminary Results and Discussion}
\begin{figure}[t]
  \centering
  \begin{subfigure}{0.4\linewidth}
    \includegraphics[width=0.9\linewidth]{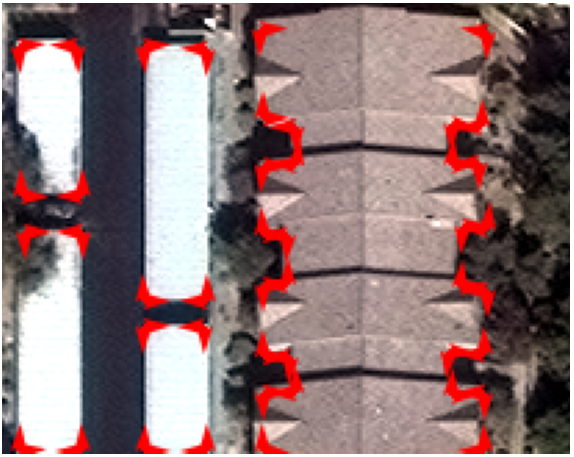}
    \caption{Prediction}
    \label{fig:orient-a}
  \end{subfigure}
  \hfill
  \begin{subfigure}{0.4\linewidth}
    \includegraphics[width=0.9\linewidth]{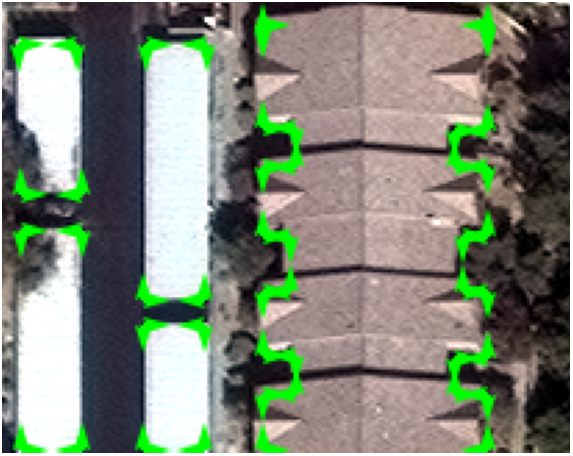}
    \caption{Ground truth}
    \label{fig:orient-b}
  \end{subfigure}
  \caption{Oriented corners on Vegas dataset. OriCornerNet can grasp building regular shapes and detect obscured corners.}
  \label{fig:orient}
\end{figure}

To assess the effectiveness of utilizing oriented corners, we conducted a comparison between the proposed method and its baseline, R-PolyGCN~\cite{rpolygcn}. Raster-based metrics, specifically AP and AR, exhibited a significant improvement of 16\% and 14\%, respectively, as shown in~\Cref{tab:vegas}. Furthermore, vector-based metrics indicated that our method achieved more precise corners without compromising polygon complexity, highlighting the effectiveness of corner-based initialization and the multi-task learning approach for orientation (see~\Cref{fig:orient}). Also, the proposed method is compared with two two-step (Mask R-CNN~\cite{maskrcnn} and FrameField~\cite{ff}) and two direct (PolyMapper~\cite{polymapper} and BiSVP~\cite{bisvp}) polygonal segmentation methods. OriCornerNet proved to outperform them by 4\% in AP and by 11\% in terms of AR.

\begin{table}
    \centering
    \caption{Quantitative results of OriCornerNet and rival methods on SpaceNet Vegas dataset. The results for the rival methods except R-PolyGCN are from \cite{bisvp}. Lower PoLiS values are better.}
    \begin{tabular}{|c|cc|cc|}
         \hline
         Method &AP & AR & PoLiS $\downarrow$ & C-IoU\\
         \hline
         Mask R-CNN~\cite{maskrcnn} & 0.47 & 0.55 & - & -\\
         FrameField~\cite{ff} & \underline{0.54} & \underline{0.59} & - & - \\
         PolyMapper~\cite{polymapper} & 0.52 & 0.58 & - & -\\
         BISVP~\cite{bisvp} & 0.53 & \underline{0.59} & - & -\\
         \hline
         R-PolyGCN~\cite{rpolygcn} & 0.43 & 0.57 & 3.20 & 0.55\\
         Ours & \textbf{0.58} & \textbf{0.70} & \textbf{2.29} & \textbf{0.80}\\
         \hline
    \end{tabular}
    \label{tab:vegas}
\end{table}

\begin{figure}[t]
  \centering
  \begin{subfigure}{0.3\linewidth}
    \includegraphics[width=0.9\linewidth]{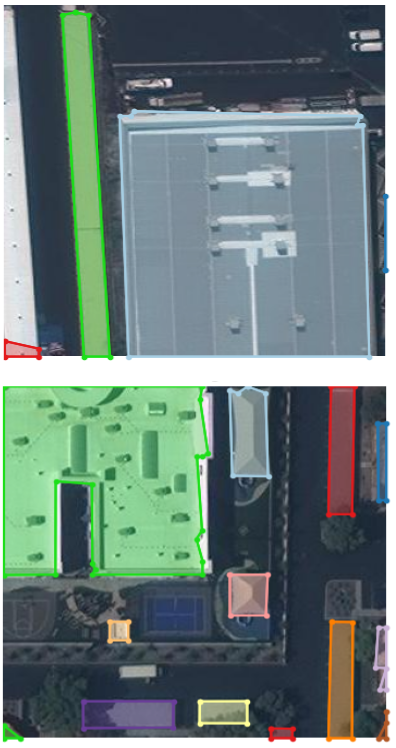}
    \caption{HiSup}
    \label{fig:crowd-a}
  \end{subfigure}
  \hfill
  \begin{subfigure}{0.3\linewidth}
    \includegraphics[width=0.9\linewidth]{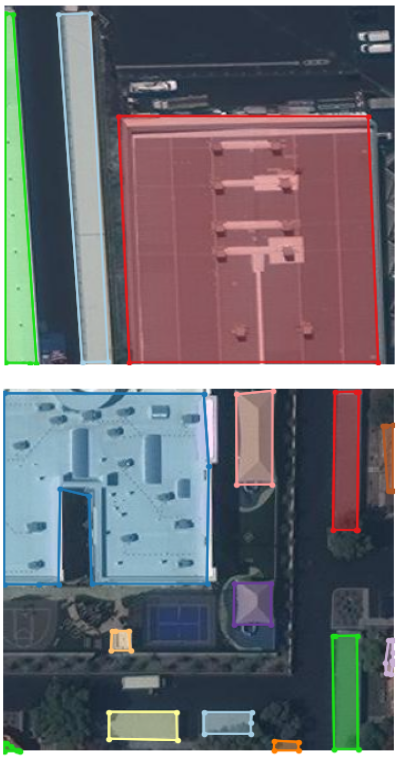}
    \caption{Ours}
    \label{fig:crowd-b}
  \end{subfigure}
  \hfill
  \begin{subfigure}{0.3\linewidth}
    \includegraphics[width=0.9\linewidth]{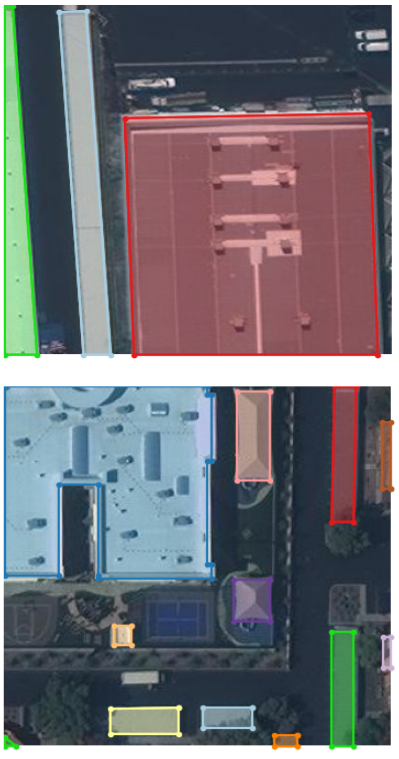}
    \caption{Ground truth}
    \label{fig:crowd-c}
  \end{subfigure}
  \caption{Qualitative results of Hisup, the proposed method, and ground truth on CrowdAI-small dataset.}
  \label{fig:crowd}
\end{figure}

An ablation study was conducted on the CrowdAI-small dataset to quantitatively evaluate the contribution of each component in the proposed method. According to~\Cref{tab:crowd}, the study revealed that combining semantic corners with the detected corners could significantly improve performance in terms of both raster-based and vector-based metrics. Additionally, the orientation consistency loss increased the AP by 0.5\% and AR by 0.2\%. Ultimately, the full method, which includes the above-mentioned components plus the orthogonality loss, delivered the best performance in terms of all metrics. Further, compared to HiSup, the state-of-the-art method, our method demonstrated competitive results in terms of AP and considerably better results in terms of AR and vector-based metrics. \Cref{fig:crowd} provides qualitative results, demonstrating the proposed method's more robust performance in predicting regular and simpler polygons.

\begin{table}
    \centering
    \caption{Ablation study of OriCornerNet and comparison to the state-of-the-art method on CrowdAI-small dataset. The lower values of PoLiS are better.}
    \resizebox{\columnwidth}{!}{\begin{tabular}{|c|cc|cc|}
         \hline
         Method & AP(\%) & AR(\%) & PoLiS $\downarrow$ & C-IoU\\
         \hline
          Baseline & 52.3 & 66.6 & 1.80 & 78.40\\
         + Semantic corners & 54.6 & 68.1 & 1.74 & 79.02\\
         + Orientation consistency loss & 55.1& 68.3 & 1.76 & 79.15\\
          + Orthogonality loss & 56.1 & \textbf{69.5} & \textbf{1.72} & \textbf{79.48} \\
          \hline
          HiSup~\cite{hisup} & \textbf{56.9} & 60.8 & 1.99 & 76.09\\
          \hline
    \end{tabular}}
    \label{tab:crowd}
\end{table}

\section{Conclusion}
In this work, we have proposed a new method for generating building polygons using oriented corners as an auxiliary representation in an end-to-end direct polygonal segmentation network. The network utilizes both semantic and geometric features to reconstruct the building polygons. It also refines them via GCNs by applying some geometric regularization terms. The experimental results on two datasets have demonstrated the effectiveness of our proposed method in extracting delineated polygons. In the future, we plan to focus on handling buildings with holes and improving the simplification using orientation information.
{
    \small
    \bibliographystyle{ieeenat_fullname}
    \bibliography{main}
}


\end{document}